\documentclass[11pt]{article}
\usepackage[utf8]{inputenc}
\usepackage{graphicx}
\usepackage{amsmath}
\usepackage{float}
\usepackage{tabularx}
\usepackage{booktabs}
\usepackage{caption}
\usepackage{hyperref}
\usepackage{geometry}
\usepackage{adjustbox}

\title{Public Machine Learning Solver Framework for Novices in the Machine Learning Domain}
\author{
  Lokman Saleh$^{1,2}$\thanks{$^{1}$ LATECE Lab, Université du Québec à Montréal, Montréal, Canada; $^{2}$ \texttt{saleh.lokman@courrier.uqam.ca}} \and
  Hafedh Mili$^{1,2}$ \and
  Mounir Boukadoum$^{1,2}$
}
\date{}

\begin{document}

\maketitle

\begin{abstract}
Solving machine learning problems is a complex process typically reserved for experts. Over the past two decades, several systems and methods have emerged to support non-experts in navigating machine learning tasks. Based on our literature review, we identify three primary categories of such tools. The first category includes fully automated systems known as AutoML (Automated Machine Learning), which use optimization techniques to automate every step of the ML workflow. The second category consists of cheat sheets created by experts; these are simple guides that help beginners choose suitable ML algorithms based on straightforward questions. The third category involves decision-support systems that rely on selection criteria—such as accuracy, transparency, and training data requirements—to guide users in prioritizing among various algorithms.

In this work, we propose a new platform that combines the strengths of the second and third categories to deliver a semi-automated, intelligent solution recommendation system for non-experts. Unlike existing approaches that only recommend a single algorithm, our platform suggests a complete pipeline tailored to the user’s problem. It integrates expert-defined selection criteria with transfer learning techniques and automatically extracts data characteristics—such as class imbalance or missing values—from user-provided datasets. The platform uses first-order logic to reason over its knowledge base and recommends suitable algorithms ranked in descending order of relevance. It features a user-friendly interface and connects to a complementary crowdsourcing platform designed specifically for machine learning experts. This connection ensures that our system remains up to date by continuously capturing new knowledge about algorithm behavior and selection criteria from experts. Moreover, the platform is built incrementally, allowing seamless integration of new algorithms, criteria, and domain knowledge over time.

To the best of our knowledge, this is the first free, publicly accessible online framework that systematically captures and operationalizes expert knowledge to guide non-experts in solving machine learning problems in a structured and transparent manner.
\end{abstract}

\textbf{Keywords:} AutoML; Selection criteria; Cheat sheet.

\section{Introduction}
\label{sec:introduction}

Machine learning is currently regarded as one of the most important scientific fields due to its powerful capabilities for knowledge discovery from large datasets. Business professionals who have accumulated data over the years increasingly leverage machine learning to gain new insights into their operations. In marketing, for example, machine learning can predict which products a customer is likely to purchase, enabling sales specialists to create personalized offers, adjust inventory levels based on predicted demand, or recommend complementary products. In the financial sector, machine learning is applied to detect fraudulent transactions and identify potential cyberattacks on banking platforms. It is also used in document processing to extract information from scanned papers using text recognition. In the medical domain, machine learning plays a crucial role in tasks such as detecting cancer through breast radiological imaging.

However, the shortage of machine learning experts presents a significant challenge in uncovering hidden knowledge from data, despite the availability of powerful algorithms. Business professionals often rely on the expertise of machine learning specialists due to the complexity of implementing and tuning these algorithms for practical applications. Access to such experts is not always readily available, and when it is, the process—from scheduling consultations to explaining business-specific problems and awaiting model evaluation—can be lengthy, sometimes taking several months.

Moreover, even for machine learning experts, identifying an appropriate solution that satisfies specific requirements can be time-consuming. Consequently, employing automated tools to assist or partially replace these experts can accelerate solution discovery, validate decisions more efficiently, and enable business professionals to gain insights from their data without the delays typically associated with expert consultations.

A concrete example highlighting the importance of automated systems is presented in \cite{RN19}. In a predictive maintenance project within the public transportation sector, a team of three data scientists worked full-time for several weeks to develop a machine learning model achieving an area under the curve (AUC) of 0.81. A few months later, the prototype of the AutoML tool DSM was applied to the same dataset—without any additional assistance or domain knowledge—to evaluate its effectiveness. Remarkably, the automatically generated model slightly outperformed the manually developed one, achieving an AUC of 0.82 after just half an hour of execution.

Bohr and Memarzadeh (2020) estimate that artificial intelligence (AI) applications, including machine learning, could reduce annual U.S. healthcare costs by up to \$150 billion by 2026. Similarly, Davenport and Kalakota (2019) anticipate that AI will soon master diagnostics and treatment recommendations in the medical field. Additionally, Gartner, a leading technology intelligence and research company, predicts that by 2020, over 80\% of all customer interactions will be managed by AI (Kumar and Trakru, 2020).

Based on our review, we identified three primary types of automated tools that assist non-experts in selecting suitable machine learning (ML) solutions. We also include a basic approach involving direct performance comparison of algorithms on a selected dataset. These four categories are:

\begin{itemize}
    \item \textit{AutoML Systems}: These systems are designed to fully automate the selection of the most appropriate ML solution based on the input dataset, without human intervention. They typically rely on optimization algorithms to explore and evaluate different models.

    \item \textit{Cheat Sheets}: Developed by large data analytics companies and based on expert knowledge, these sheets present decision trees or printed guidelines that help users choose suitable ML algorithms for a given problem.

    \item \textit{Selection Criteria Systems}: These systems offer an interactive approach, allowing users to define specific criteria or problem characteristics. Based on this input, the system suggests appropriate algorithms. The design of such systems generally requires substantial expert knowledge.

    \item \textit{ML Algorithm Comparison}: This naive approach involves manually testing and comparing multiple machine learning algorithms on a given dataset to determine which performs best. Unlike the previous categories, it does not rely on expert systems or automation. Given that this method has been thoroughly addressed in existing surveys~\cite{Andreopoulos2009, Das2017, Fatima2017, Patil2014, Saxena2017}, we do not cover it in detail in this study.
\end{itemize}

Our objective is to develop an automated system for selecting machine learning solutions that leverages the strengths of existing tools while addressing their key limitations.

Cheat sheets developed by reputable organizations such as Microsoft, Scikit-learn, and SAS serve as valuable resources for guiding beginners in choosing suitable ML algorithms. These tools rely on fundamental criteria including data type, dataset size, interpretability, accuracy requirements, and the nature of the learning task (i.e., supervised vs. unsupervised). However, cheat sheets often adopt a tree-like structure, which limits their scalability and flexibility. Additionally, they generally lack detailed guidance on crucial preprocessing steps that can significantly affect the performance of the selected algorithm.

\begin{figure}[H]
    \centering
    \includegraphics[width=\textwidth]{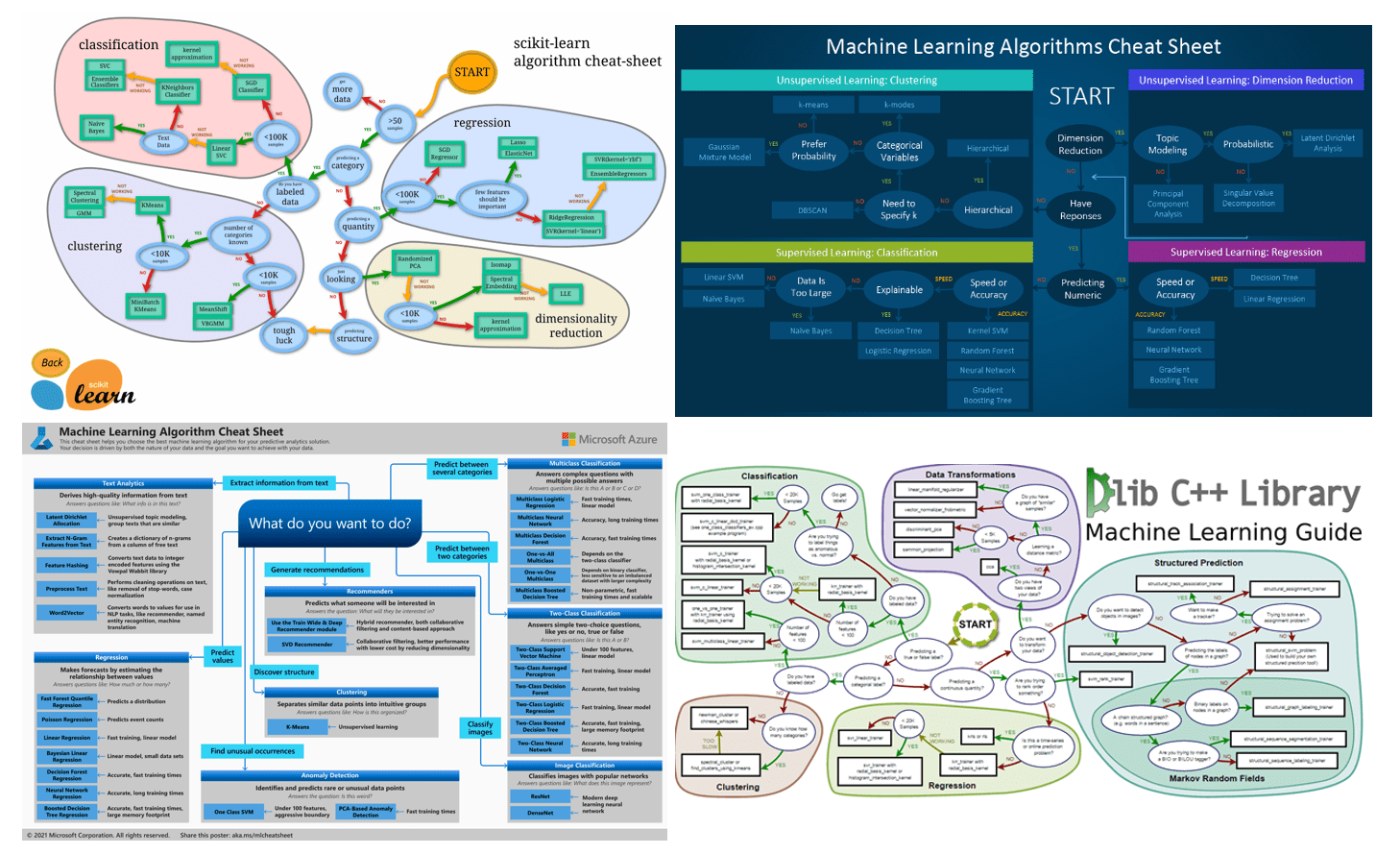}
    \caption{The cheat sheets from Microsoft, SAS, and Scikit-learn.}
    \label{fig:Cheat-sheet}
\end{figure}

Selection criteria systems aim to identify the most appropriate ML algorithm based on dataset characteristics and user-defined requirements. While promising in concept, existing systems~\cite{RN50, RN49, RN54} remain limited in practice. These approaches typically rely on the expertise of a small number of ML specialists—often the authors of the respective studies—and support only a single algorithm recommendation without accommodating full ML pipelines. Furthermore, knowledge about algorithm performance (e.g., comparative accuracy between algorithms) is often described narratively in research articles, with no intelligent mechanism for processing or applying this information in practice.

In the case of automated support-focused studies (AutoML systems), many of the proposed tools or frameworks are still incomplete, difficult to use, or require substantial expertise to interpret and apply effectively. These systems often fail to consider critical factors or selection criteria and are typically limited to solving relatively simple machine learning problems~\cite{RN54, RN50, RN73}. Moreover, some recent research, such as~\cite{RN49}, introduces methods that use machine learning to select other machine learning algorithms. However, such approaches often lack comprehensive explanation, validation, and robustness, and therefore fall short of realizing full automation.

According to the standard definition, a fully functional AutoML system should satisfy the following key criteria:

\begin{enumerate}
    \item No human intervention required.
    \item Rapid execution and response time.
    \item Capability to handle a wide range of machine learning problems across different domains.
\end{enumerate}

However, our review of existing AutoML systems reveals that, although they successfully eliminate the need for human intervention, they still encounter several significant challenges:

\begin{enumerate}
    \item \textbf{Slow execution:} These systems can be time-consuming, often requiring extended periods to identify an optimal solution.
    \item \textbf{Limited task coverage:} Most AutoML tools primarily focus on classification and regression tasks, leaving out a broader range of machine learning problems such as clustering, anomaly detection, or time-series forecasting.
    \item \textbf{Narrow evaluation criteria:} They frequently rely on a single performance metric—such as accuracy—which may be insufficient for comprehensive evaluation, particularly in imbalanced or multi-objective scenarios.
    \item \textbf{Lack of transparency:} Many AutoML systems operate as black boxes, offering little insight into the reasoning behind their model selection or optimization strategies.
    \item \textbf{Limited flexibility and extensibility:} These systems often do not support the integration of new evaluation criteria or emerging learning algorithms, reducing their adaptability to evolving user needs or research advances.
\end{enumerate}

Therefore, further improvements are needed for AutoML systems to fully satisfy the criteria outlined in their definition.

In this work, we present a novel automated ML solution selection system, available online at \texttt{isolvemymlproblem.org}. This system is designed to assist users—particularly non-experts—in identifying and constructing suitable ML pipelines based on user-defined criteria (e.g., accuracy, transparency, data distribution) and problem descriptions.

The system accepts textual descriptions of the ML problem, which are processed using transfer learning techniques based on transformer architectures. Users can also connect to their own MySQL servers to retrieve datasets, select relevant attributes, and automatically generate a new dataset with computed characteristics. These characteristics include attribute variance, value distribution, presence of missing values, number of features, dataset size, attribute correlations, and more. Based on this information, the system identifies the learning type (e.g., supervised, unsupervised, classification, regression, or clustering).

Using both the user’s textual problem description and the extracted data characteristics, the system proposes a candidate ML pipeline. This pipeline is then refined using a first-order logic engine that incorporates best practices in the field. For example, if the selected algorithm is ID3, the engine recommends discretizing numerical attributes; if the dataset is imbalanced with a very small minority class, it suggests downsampling.

The pipeline’s model construction component presents a ranked list of ML algorithms, ordered by relevance. This ranking is based on two key factors: (i) selection criteria extracted from the dataset and specified by the user, and (ii) expert-sourced selection criteria collected through the platform \texttt{icontributetoml.org}, where ML practitioners share knowledge about algorithm suitability under various conditions.

Ultimately, the user receives a ready-to-execute ML pipeline tailored to their specific problem. This pipeline is represented using Business Process Model and Notation (BPMN), offering a clear and structured visualization of each step.

Currently, the system does not support direct execution of the generated pipeline due to incomplete implementation of execution scripts for all referenced algorithms. A detailed evaluation comparing the system's recommendations with manual selections made by experienced ML experts will be presented in future work. Nevertheless, the framework provides a comprehensive, structured pipeline as a string output, which users can execute independently on their datasets.

\section{State of the Art}
\label{Satet-of-the-arts}

Several research studies have aimed at listing, understanding, analyzing (advantages and disadvantages), and comparing existing algorithms \cite{RN48, RN58, RN83, RN56, RN62, RN80}. On the other hand, few works have focused on developing tools to assist with algorithm selection \cite{RN51, RN53, RN54, RN81, RN50, RN49, RN52, RN73}. Studies that directly answer our first research question, and those that offer ideas, architectures, or processes for selecting the right ML algorithm, include:

Bermudez tested a set of classification algorithms to select the correct classifier \cite{RN49}. Sala et al. \cite{RN50} proposed a simple framework composed of two layers: the first reduces the number of algorithms according to their training scope (classification, regression, or clustering), and the second provides four selection criteria to guide the user’s final choice. Some studies (e.g., Microsoft, 2020; Scikit-learn, 2020) \cite{RN92, RN55} focused on one or two selection criteria. Microsoft (2020) \cite{RN92} used training time and model accuracy (prediction, clustering, regression, etc.) as the main factors for selecting the right learning algorithm. Scikit-learn (2020) \cite{RN55} focused on data size as the main criterion. These two studies proposed guides (cheat sheets) to direct an expert toward the right choice of algorithm. Metwalli (2020) \cite{RN47} divided the criteria for selecting a learning algorithm into two parts: i) those related to the learning base (size, type, etc.) and ii) those related to user demand (accuracy, training time, etc.). Other approaches such as Akaike (1974) \cite{RN51}, Hannan and Quinn (1979) \cite{RN53}, and Schwarz (1978) \cite{RN52} use model quality measures like AIC, BIC, or HQC, which penalize the algorithm according to the number of parameters to adjust. Kotsiantis et al. \cite{RN54} identified thirteen selection criteria for six algorithms and assigned a score from one to four for the correlation between each characteristic and algorithm. Fatima and Pasha \cite{RN56} attempted to find the algorithm with the highest accuracy for each of five diseases (cardiac, diabetes, etc.), assuming that the best algorithm will always be associated with the same problem. Andreopoulos et al. \cite{RN48} attempted to find important, desirable, or required selection criteria for clustering algorithms in the biomedical domain, without proposing a process to guide the reader toward the right choice of algorithm.

Since 2013–2015, the research community has started to adopt fully automatic solutions based on optimizers. These systems automatically select the best approach in terms of precision or accuracy for a specific ML problem, without requiring the intervention of ML experts. Some systems, like AutoStacker \cite{RN30}, are specialized for handling small datasets, while others, such as D-Smart \cite{RN2} and H2O \cite{RN28}, are designed to manage large datasets with support for distributed solutions. Additionally, AutoSklearn \cite{RN32} (with KNN), Smart-ML \cite{RN11} (using gradient boosting), meta-learning approaches, and Zooming \cite{RN17} make use of meta-learning techniques to establish a data knowledge base initialized with meta-features.

Beyond meta-feature values, this data knowledge may also incorporate information such as training times (tunability) and algorithm accuracy. Across these AutoML systems, the primary pipeline components typically include data preprocessing, feature selection and extraction, machine learning algorithms, and ensemble learning. Moreover, most systems are designed for either classification or regression tasks.

Some systems, such as H2O \cite{RN28} and AutoGluon \cite{RN29}, rank ML algorithms based on input from ML experts to initiate optimization with the most promising option. For H2O, users need to specify relevant hyperparameters and their value ranges using a random optimizer. In the case of PSPSO \cite{RN1}, users must also specify relevant hyperparameters, with only continuous values being applicable.

The choice of optimizer varies: AutoWeka \cite{RN108} and AutoSklearn \cite{RN32} use Bayesian optimization, while ML-Plan \cite{RN12} and RECIPE \cite{RN22} employ genetic algorithms with specific approaches. PBIL \cite{RN239} uses a genetic algorithm with pipelines consisting of three components (primitive ML algorithms, ensemble learning, and hyperparameter optimization), each weighted accordingly. In DarwinML \cite{RN245}, individuals are represented as graphs, with edges modified using a probability matrix.

MOSAIC \cite{RN262} optimizes pipeline structure with MCTS and hyperparameters using Bayesian optimization. In the POSH \cite{RN236} system, a portfolio of predefined pipelines is employed, and Bayesian optimization is combined with SH (Successive Halving) algorithms to provide additional training data for the most promising pipelines.

It is noteworthy that some AutoML systems like P4ML \cite{RN237} are versatile and capable of handling non-tabular datasets, such as images and audio. For ensemble learning, AutoSklearn \cite{RN32}, AutoStacker \cite{RN30}, and Gluon \cite{RN29} use majority voting, while H2O \cite{RN28} leverages the SuperLearner algorithm.

\section{I Solve My ML Problem}
\label{I-solve-my-ml-problem}

In our previous work, which focused on the selection of machine learning algorithms, we introduced a platform designed to aggregate expert knowledge in ML, available at \url{http://icontributetoml.org/help}. In the present study, we build upon that foundation by presenting the theoretical framework that enables the integration of expert knowledge into the algorithm selection process. We also discuss the selection criteria used to adapt algorithm choices based on user needs and dataset characteristics, and we provide an overview of the different categories of learning algorithms considered.

In the section on pipeline construction, we detail the process of generating and refining ML pipelines. Our approach consists of two key stages: (i) mapping the user's textual problem description to a suitable pipeline structure using transfer learning, and (ii) refining the selected pipeline using a set of rule-based transformations. These rules—validated and approved by the research community—allow the addition or removal of components to tailor the pipeline to the specific problem context.

Finally, we emphasize that the pipeline is dynamically influenced by the selected learning algorithm. Therefore, modifications (e.g., adding preprocessing steps) may be necessary to ensure compatibility. For instance, decision tree algorithms typically require discretized input data. Thus, if the user selects a decision tree while working with continuous data, the system will recommend or apply a discretization step to ensure correct execution. This bidirectional interaction between pipeline generation and algorithm selection enhances the adaptability and correctness of the final ML solution.

\subsection{Selection Criteria}

\textit{Definition — What Are Selection Criteria?}

Selection criteria refer to a set of characteristics that describe and differentiate machine learning algorithms. These criteria help identify each algorithm's capabilities (e.g., classification, regression, clustering), strengths and weaknesses (e.g., susceptibility to overfitting, sensitivity to outliers, handling of missing data), input-related aspects (such as data distribution, dataset size, feature dimensionality, and required hyperparameters), and output-related considerations (such as accuracy, explainability, interpretability, and training time).

\textit{Advantages of Our Approach Using Selection Criteria}:

\begin{itemize}
    \item \textbf{Fast:} Our approach avoids time-consuming optimization processes typically used in AutoML.
    \item \textbf{Explicit:} The decision-making process is transparent, enabling domain experts to understand why a particular algorithm was selected.
    \item \textbf{Controlled:} Users retain full control over the final solution, allowing manual adjustments if needed.
    \item \textbf{Professional:} Even if expert knowledge does not directly lead to the optimal solution, it can still guide the search in a meaningful and informed direction.
    \item \textbf{Interpretable by ML Experts:} The system is designed to be understandable and modifiable by ML professionals, who can refine, extend, or remove selection criteria as needed.
    \item \textbf{Performant:} The criteria-driven process helps determine appropriate preprocessing and feature engineering steps, improving pipeline quality.
    \item \textbf{Diversifiable:} Unlike many AutoML solutions limited to classification or regression, our approach can support a wider variety of ML tasks without significantly increasing computational complexity.
    \item \textbf{Organized:} Recommended ML algorithms are ranked according to their relevance to the user's specific problem.
    \item \textbf{Open:} The platform is open to contributions from the broader ML community, enabling continuous enrichment of the underlying knowledge base.
\end{itemize}

\subsection{ML Algorithm Selection Process}

\begin{figure}[H]
    \centering
    \includegraphics[width=\textwidth]{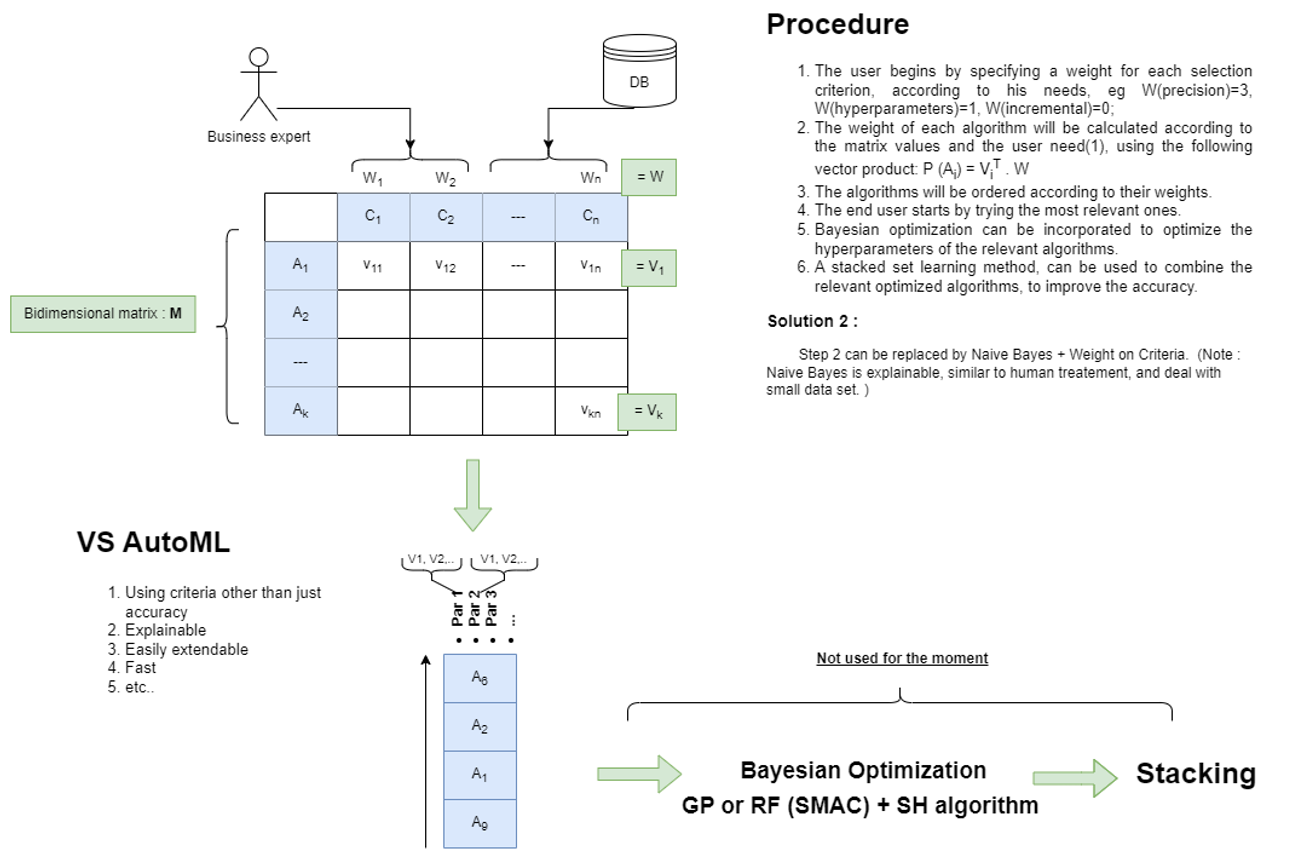}
    \caption{Machine learning algorithm selection procedure.}
    \label{fig:image4.13}
\end{figure}

\textbf{Selection Procedure:}

\begin{enumerate}
    \item The domain expert starts by assigning weights to each selection criterion based on their requirements. For example, they may choose \(W(\text{accuracy}) = 3\), \(W(\text{interpretability}) = 1\), \(W(\text{incrementality}) = 0\), and \(W(\text{distributability}) = 0\). Simultaneously, weights are automatically assigned to criteria related to the training dataset, such as \(W(\text{imbalanced data}) = 3\) and \(W(\text{high dimensionality}) = 2\), etc.
    
    On the other hand, we have the bidirectional matrix \(M\) containing the knowledge of ML experts in numerical form. For example, \(V_{11}\) in the matrix may contain the value "3", which corresponds to the average weight assigned by ML experts regarding the accuracy criterion for the Naive Bayes algorithm.
    
    \item The weight of each algorithm is calculated based on the values of the ML experts matrix \(M\) and the weight vector \(\vec{W}\) filled by the domain expert (step 1), using the Cartesian product. For example, to calculate the weight of algorithm \(A_i\), \(P(A_i) = \vec{V_i^T} \cdot \vec{W}\) (see Figure~\ref{fig:image4.13}). Similarly, the weights of all algorithms are given by \(\vec{P(A)} = M \cdot \vec{W}\). Based on the values of \(\vec{P(A)}\), the algorithms are ranked. Thus, the algorithm with the highest value is prioritized for the domain expert to use in solving their ML problem.
    
    \begin{itemize}
        \item \textbf{Similarity Measures}: In addition to the Cartesian product, other methods such as Euclidean distance or cosine similarity can be used for ranking. These measures compare the criterion values of each algorithm with those specified by the user, thus evaluating their proximity. Algorithms are then ranked based on this proximity to the user's specific needs.
        
        The distinction between the Cartesian product and similarity measures is that the Cartesian product provides a general solution based on criteria defined by the data and domain experts, while similarity measures aim to find a more precise solution based on the specified criteria. For example, if accuracy specified by the domain expert has weight 2 and an algorithm has weight 5 for the same criterion, the Cartesian product yields \(2\times5=10\) for that algorithm. For another algorithm with weight 7, the result is 14. The Cartesian product favors the second algorithm due to its overall reliability, even if it deviates from the user's specification. In contrast, distance measures would choose the algorithm with weight 5 as being closer to the user's need.
        
        \item \textbf{Naive Bayes}: Additionally, Naive Bayes can be used. Instead of using a matrix containing the average response for each cell, we use all the data collected by ML experts as training data. Naive Bayes is motivated by its ability to explain results in a manner similar to human reasoning, its effective handling of small datasets, and the independence of our selection criteria. The posterior probability is:
        \[
        P(A \mid C_1, C_2, C_3) = P(C_1 = W(C_1) \mid A) \times P(C_2 = W(C_2) \mid A) \times P(C_3 = W(C_3) \mid A) \times P(A)
        \]
        where \(C_i\) is a selection criterion and \(W(C_i)\) is the weight assigned by the domain expert. Here, \(P(C_i \mid A)\) could be replaced by a normal distribution, using the mean and variance of responses for that criterion and algorithm. Alternatively, discretization methods such as equal width, equal frequency, entropy-based, or supervised discretization can be applied \cite{fayyad1993multi}.
        
        Currently, the Bayesian approach is not integrated into our platform because our knowledge base has not yet been enriched by many ML experts. This idea is novel and will be implemented during the pipeline execution stage.
    \end{itemize}
    
    \item The algorithms are ranked according to their weights.
    \item The end user starts by trying the most relevant ones.
    \item Bayesian optimization will be integrated to optimize relevant algorithms' hyperparameters using values collected from ML experts.
    \item A stacked ensemble learning method can also be used to combine relevant optimized algorithms to improve accuracy.
\end{enumerate}

In the approach described above, Bayesian optimization will be employed for hyperparameter tuning. On our website, ML experts can specify which hyperparameters are relevant for each ML algorithm and the values typically used for them (see \url{http://icontributetoml.herokuapp.com/help}). This targeted approach—focusing on relevant configurations rather than optimizing all hyperparameters—has shown promising results \cite{VanRijn2018, Feurer2018Eggensperger} and aims to minimize the search space, where Bayesian optimization excels \cite{Kotthoff2019}. Additionally, a successive halving algorithm will allocate more training time to the best-performing algorithm for the current task.

Finally, stacked ensemble learning will be employed with the best algorithm configurations to improve accuracy. This method has proven effective, especially with small training datasets \cite{ChenBoy2018, Erickson2020}, which is significant because acquiring large training datasets is often a major challenge for domain experts.

In our platform, only the first three points of this process have been implemented and validated. Points 4 through 6 will be explored and validated during later stages of our research project.

\subsection{Process of Selecting and Constructing a Pipeline}

The pipeline construction process consists of two parts:

\begin{enumerate}
    \item \textbf{Selection of a processing chain from our portfolio}. The goal is to compare the semantic similarity of the business specialist's problem description with the description of each chain in our portfolio. This is achieved with high precision using pre-trained deep neural networks (transfer learning) that solve general problems without needing to readjust network weights. The pre-trained algorithm used is Sentence-BERT \cite{RN308}, which returns a score measuring semantic similarity between two texts. We tested it extensively, and it is able to find the right chain even with a simple problem description.

    \item \textbf{Refinement of the selected chain} to better suit the problem using a rule base built from selection criteria representing the database and the business expert's requests. These rules are managed using a rule base management engine based on first-order logic, implemented with Microsoft's Z3 engine in Python. First-order logic relies on the resolution rule \cite{RN365}:
    \[
    A \cup B \text{ and } \neg B \cup Y \quad \text{implies} \quad A \cup Y
    \]
    For example, if \(A \cup B\) is true and \(\neg B\) is true, then \(A\) is true. Rules are scalable because they are stored in a Python file, allowing dynamic modification on the server side without redeploying the project.
\end{enumerate}

\subsection{How “I Solve My ML Problem” Works}
\label{I-solve-my-ml-problem-how-it-work}

The platform integrates and implements the algorithm and pipeline selection and refinement processes described above. It is intended for ML novices or business experts to solve ML problems without requiring expert ML skills. The platform consists of four parts:

\begin{itemize}
    \item \textbf{Problem Domain}: The user describes their problem and specifies selection criteria.
    \item \textbf{Data Characteristics}: The platform retrieves the database and automatically determines data-related selection criteria.
    \item \textbf{Data Analysis Problem}: The type of ML problem to solve is specified.
    \item \textbf{Draft Solution}: The proposed pipeline is presented, with algorithms chosen for each component.
\end{itemize}

The platform is available at \url{http://isolvemymlproblem.org} and was implemented using HTML5, CSS3, Bootstrap, jQuery, MySQL, JavaScript, Node.js, React, Python, Spring Boot, REST, etc.

\subsubsection{Problem Domain}
\label{domain-problem}

\begin{figure}[H]
    \centering
    \includegraphics[width=0.5\textwidth, angle=270]{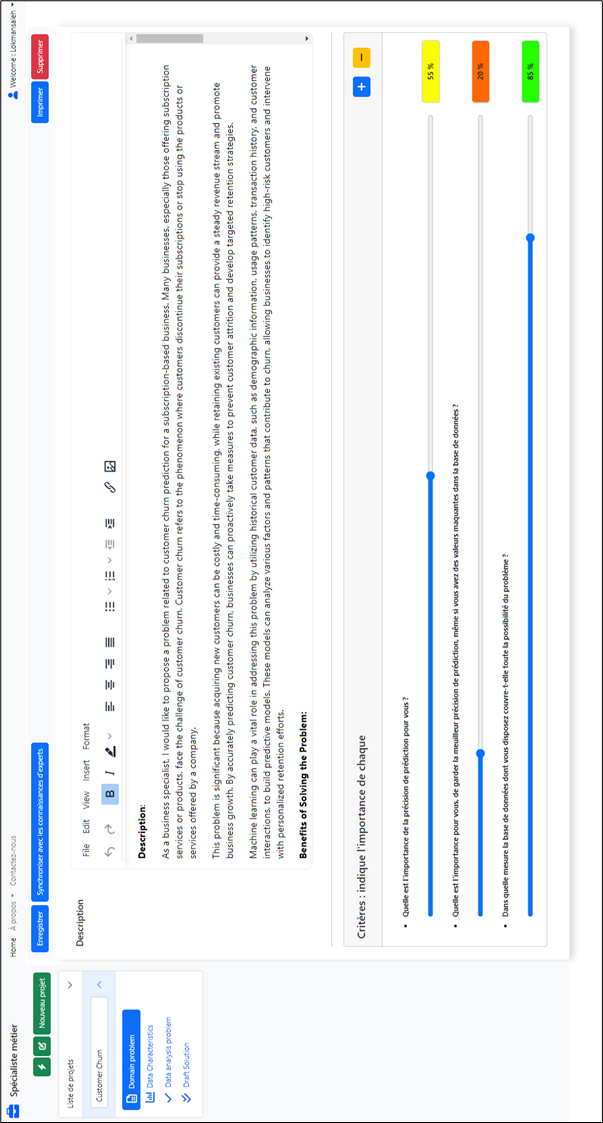}
    \caption{Domain problem page.}
    \label{fig:isolve-home-page}
\end{figure}

On this page, the user specifies two elements:

\begin{enumerate}
    \item \textbf{Problem description} (see Figure~\ref{fig:isolve-home-page}).
    \item \textbf{Selection criteria}: The user indicates the importance of each criterion relative to their context (see Figure~\ref{fig:isolve-home-page}).
\end{enumerate}

\subsubsection{Data Characteristics}

On the data characteristics page, the user must grant the platform access to their database. We offer two solutions:

\begin{itemize}
    \item Upload an Excel file.
    \item Connect to a relational database.
\end{itemize}

\begin{figure}[H]
    \centering
    \includegraphics[width=\textwidth]{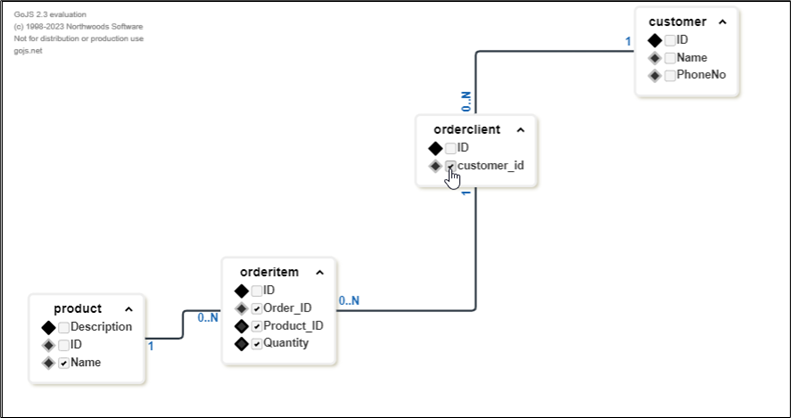}
    \caption{Database schema.}
    \label{fig:Data-base-Schema}
\end{figure}

We focus on the second solution as it is most common in businesses. To connect to a remote relational database, the user enters the server authentication data (database name, user name, password). The platform then retrieves the database schema (Figure~\ref{fig:Data-base-Schema}), from which the user selects relevant attributes.

\begin{figure}[H]
    \centering
    \includegraphics[width=0.6\textwidth]{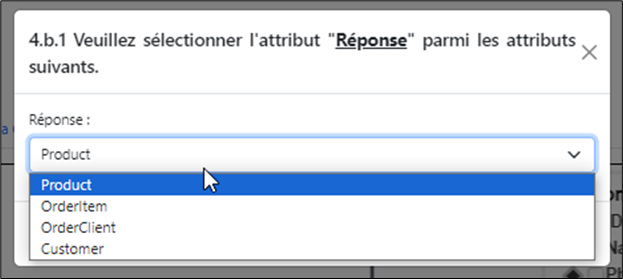}
    \caption{Response attributes.}
    \label{fig:Response attributs}
\end{figure}

Next, the user specifies whether there is an attribute that is the answer (class attribute) to be predicted. This is done using a drop-down list (Figure~\ref{fig:Response attributs}).

By clicking “Create query”, the system automatically generates an SQL query that joins the selected attributes. A sample of 100 records, database features, and data-related selection criteria are displayed in separate tabs.

\subsubsection{Data Analysis Problem}
\label{data-analysis-problem}

The problem type (e.g., supervised vs. unsupervised) is automatically determined using data criteria from the previous step and user requests from the problem domain. The user can modify this choice if needed.

\begin{figure}[H]
    \centering
    \includegraphics[width=\textwidth]{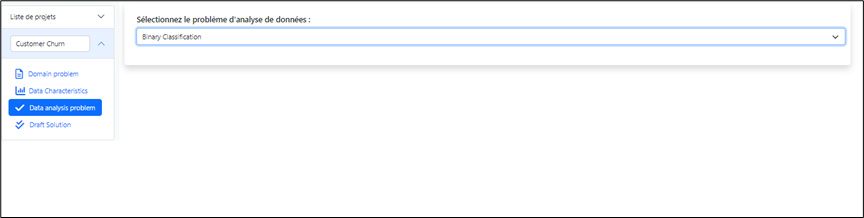}
    \caption{Data analysis problem page.}
    \label{fig:data analysis problem}
\end{figure}

\subsubsection{Draft Solution}
\label{draft-solution}

Based on all the information entered, automatically extracted, or deduced, the processing chain is selected, reconstructed, or improved, including the choice of learning algorithm in the “Build model” component.

\begin{figure}[H]
    \centering
    \includegraphics[width=\textwidth]{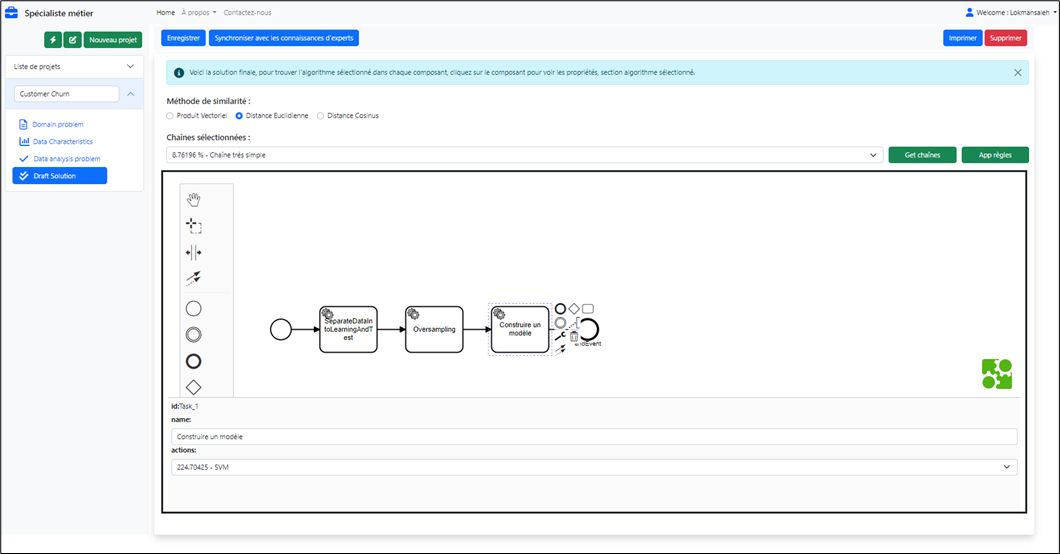}
    \caption{Draft solution page.}
    \label{fig:draft solution}
\end{figure}

The two platforms (\texttt{icontributetoml.org} and \texttt{isolvemymlproblem.org}) share three tables: users, selection criteria, and learning algorithms. This allows them to benefit from each other and synchronize data, making our solution scalable. New selection criteria or algorithms added by experts become immediately available.

Synchronization between the platforms is performed by the website administrator using the “Sync with expert knowledge” button. The user can also choose among selection methods (dot product, Euclidean distance, cosine distance) as shown in Figure~\ref{fig:draft solution}. Changing the method or modifying selection criteria on previous pages automatically and instantly recalculates algorithm relevance without page refresh.

For pipeline selection, a “Get Chains” button retrieves relevant chains. The most relevant chain according to the transformer algorithm receives the highest weight. Applying rules is optional; the user can click “Apply Rules” to refine the chosen chain. Furthermore, changing the ML algorithm in the chain may require new preprocessing components (e.g., discretization for ID3). Both algorithm and pipeline selection are dynamic and instantaneous. The system is fully scalable: adding a selection criterion, algorithm, rule, or chain is immediately reflected without redeployment.

\subsection{Navigation Bar}

Each connected client can create multiple projects using the left navigation bar, switch between projects, change project names, save changes, print, delete projects, etc. Most actions are confirmed via modal pop-ups.

\subsection{Validation}

To validate the selection process, we filled in selection criteria values on the domain expert platform to match those filled in by ML experts on \texttt{icontributetoml.org} for a specific learning algorithm. If the same algorithm is preferred, the process is valid.

\begin{enumerate}
    \item \textbf{Naive Bayes}: The selection criteria values for Naive Bayes filled in by an ML expert were used. The domain expert platform selected Naive Bayes as the best among proposed algorithms using dot product, cosine distance, and Euclidean distance. This confirms that the selection process works well (see Figure~\ref{fig:image7.16}).

    \begin{figure}[H]
        \centering
        \includegraphics[width=\textwidth]{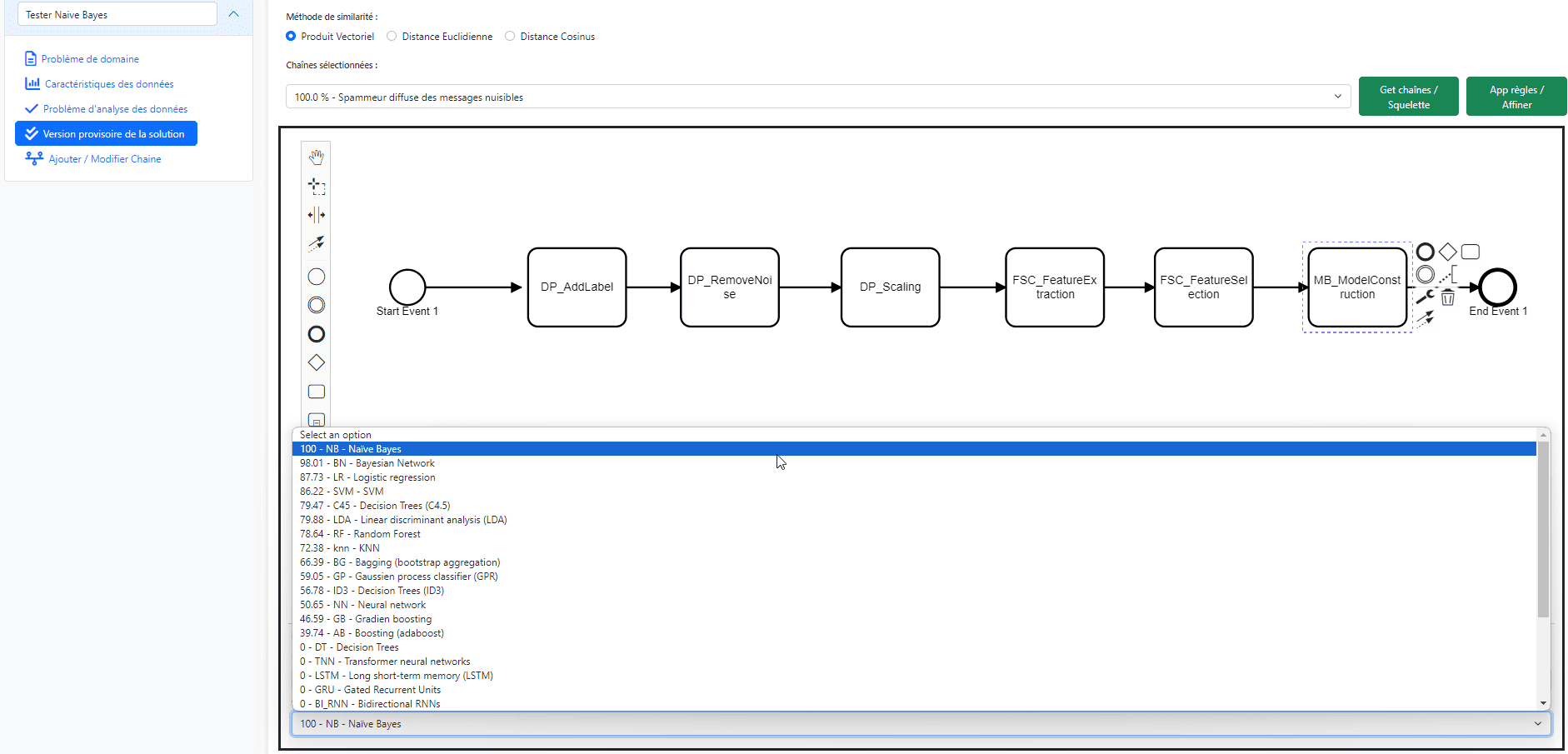}
        \caption{Naive Bayes obtained the highest score using the dot product.}
        \label{fig:image7.16}
    \end{figure}

    \item \textbf{KNN}: The same test was repeated for KNN, and KNN achieved the highest score.
\end{enumerate}

These figures demonstrate that our system can recommend the ML algorithm that meets the domain expert's requirements and dataset characteristics. We repeated the test for Bayesian Network, Random Forest, Neural Network, and SVM, and the system correctly recommended the algorithm 100\% of the time (out of the six algorithms tested). This result indicates that if the selection criteria values provided by ML experts are correct, the system will propose the correct algorithm with 100\% accuracy (using dot product, Euclidean distance, or cosine measure).

\section{Conclusion}

In this work, we have presented our final platform, which integrates all the processes, algorithms, knowledge, and skills described in previous sections. To solve an ML problem using our platform, a domain specialist follows these steps:

\begin{enumerate}
    \item Describe the problem and select the criteria values on the “Problem Domain” page.
    \item Specify server information and the database to be used as the training base on the “Data Characteristics” page. Select relevant attributes; an SQL query is automatically generated. The user can view and manually correct database characteristics if necessary.
    \item On the “Data Analysis Problem” page, the platform automatically determines the learning type from the database. The user can manually modify it if needed.
    \item On the “Draft Solution” page, the user visualizes the proposed processing chain and the algorithms for each component. A separate “Add, Modify a Processing Chain” page allows ML experts to add, modify, or delete processing chains, algorithms, components, and selection criteria.
\end{enumerate}

Our platform has demonstrated 100\% precision in selecting the correct machine learning algorithm and high accuracy in selecting the appropriate pipeline based on the problem description. However, refining the selected pipeline and choosing the right algorithm for each pipeline component are not yet performing optimally due to the limited size of the expert knowledge base. Nevertheless, the results obtained are consistent with the knowledge stored in the database.

To our knowledge, our system is innovative in its approach to solving the ML solution selection problem. This thesis presents advancements at different levels: problem translation, capturing domain expert requirements via questions, pipeline description language, pipeline templates, rule base, pipeline selection and refinement processes, component algorithm selection, and the two platforms for domain experts and ML experts. Although existing AutoML systems are competitive in prediction accuracy, our approach offers several advantages, including instant response, consideration of domain expert requests, explainability, and more. Additionally, based on our thorough study of existing AutoML systems, we have outlined a new AutoML system (currently theoretical).

Future work includes: (i) implementing and validating the proposed optimizer-based AutoML system, (ii) enriching our knowledge base, (iii) executing the processing pipeline, and (iv) inviting domain experts to test our platform at \url{http://isolvemymlproblem.org}.

\bibliographystyle{plain}
\bibliography{bibliography/references}

\end{document}